\documentclass{article}
\usepackage[utf8]{inputenc} 
\usepackage{spconf,amsmath,graphicx,color}
\usepackage{amsmath}
\usepackage{amsfonts}
\usepackage{amssymb}
\usepackage{graphicx}
\usepackage{blindtext}
\usepackage{float}
\usepackage[margin=0.8in]{geometry}
\usepackage{url}
\usepackage{color}
\usepackage{lipsum}
\usepackage{mathtools, nccmath}

\title{Always Look on the Bright Side of the Field: Merging Pose and Contextual Data to Estimate Orientation of Soccer Players}
\name{A. Arbués-Sangüesa$^{1}$, A. Martín$^{1}$,  J. Fernández$^{2}$, C. Rodríguez$^{2}$, G. Haro$^{1}$, C. Ballester$^{1}$}
\address{$^{1}$Universitat Pompeu Fabra, $^{2}$ Futbol Club Barcelona}
\begin{document}
%

\maketitle
\begin{abstract}
Although orientation has proven to be a key skill of soccer players in order to succeed in a broad spectrum of plays, body orientation is a yet-little-explored area in sports analytics' research. Despite being an inherently ambiguous concept, player orientation can be defined as the projection (2D) of the normal vector placed in the center of the upper-torso of players (3D). This research presents a novel technique to obtain player orientation from monocular video recordings by mapping pose parts (shoulders and hips) in a 2D field by combining OpenPose with a super-resolution network, and merging the obtained estimation with contextual information (ball position). Results have been validated with players-held EPTS devices, obtaining a median error of 27 degrees/player. Moreover, three novel types of orientation maps are proposed in order to make raw orientation data easy to visualize and understand, thus allowing further analysis at team- or player-level. 

\end{abstract}


\section{Introduction}
\label{sec:intro}
\begin{figure}[!t]
\includegraphics[width=0.5\textwidth]{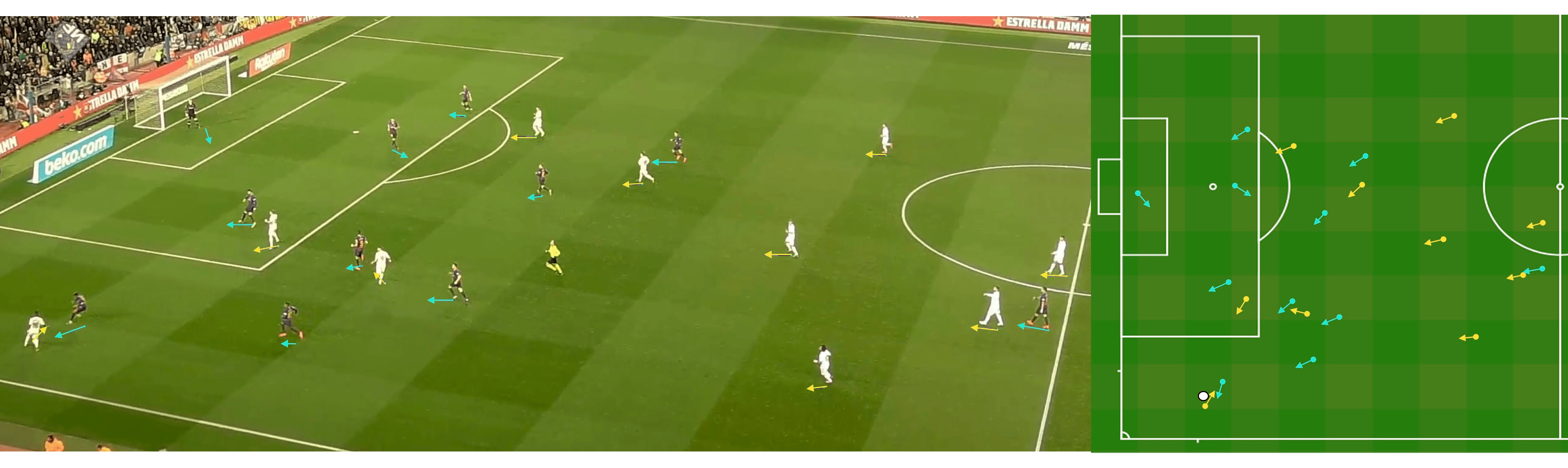}
 \caption{Qualitative results, displayed in the original frames (left), and the 2D field (right).}
  \label{fig:VisRes}
\end{figure}

The recent rise of sports analytics has provided a new set of metrics and statistics that can serve coaches to evaluate both player's and team performance. From spatio-temporal models that estimate the probability of possession success in soccer \cite{fernandez2019decomposing}, to the forecasting of future movement locations in basketball \cite{seidl2018bhostgusters}, tracking data has provided a rich source of information for exploring complex spatio-temporal dynamics in team sports (\textit{e.g., }\cite{gao2019graph,girdhar2018detect,jin2019multi,manafifard2017survey,ran2019robust,wang2019unsupervised,wang2019learning}). Despite their importance, these location data are clearly insufficient to determine if a player is in condition of properly acting during the play, which will be influenced by contextual information and the player's own pose and orientation. Proper orientation has proven to be crucial for soccer players in order to excel in particular situations, such as receiving or giving ideal passes because of an appropriate field of view, defending two players at a time or finding open-spaces due to a fast reaction. Body orientation is claimed to be more meaningful in the sports' context than gaze orientation (given by methods such as  \cite{kellnhofer18gaze360,fischer2018rt}); nevertheless, only few contributions have been made about body orientation in sports' challenging scenarios  \cite{felsen2017body,fastovets2013athlete}. The main goal of this article is to estimate the body orientation of soccer players from video data, with potential generalization to other sports. By seeking the 2D orientation of the field projection of the normal vector placed in the center of the upper-torso of players, this paper presents a novel technique to extract orientation by merging pose and contextual information. On the one hand, OpenPose \cite{openpose} is used in combination with a super-resolution network \cite{zhang2018residual} to extract the coordinates of body parts of every single player; projectively mapping key pose parts (in particular, shoulders and hips) in a 2D field-space results in a first orientation estimation; on the other hand, contextual information quantifies the orientation of each player with respect to the ball. The interaction between the ball and the players has been acknowledged as important for action analysis (e.g., \cite{maksai2016players,felsen2017will,kamble2019ball,thomas2017computer}). Results have been obtained by validating the output of the presented method with data extracted from players-held EPTS \cite{fifaEpts} devices: 96.5\% accuracy on left-right orientation directions is obtained together with an absolute median error of 27.66 degrees/player; a visual example can be seen in Fig.~\ref{fig:VisRes}. Besides, as raw orientation data might be difficult to interpret, three different visualization tools are proposed to analyze the orientation information in relation with different events or match context: OrientSonar, Reaction, and On-field maps. 

\section{Proposed Method}
\label{sec:methods}
\begin{figure*}[h!]
\centering
  \includegraphics[width=0.8\textwidth]{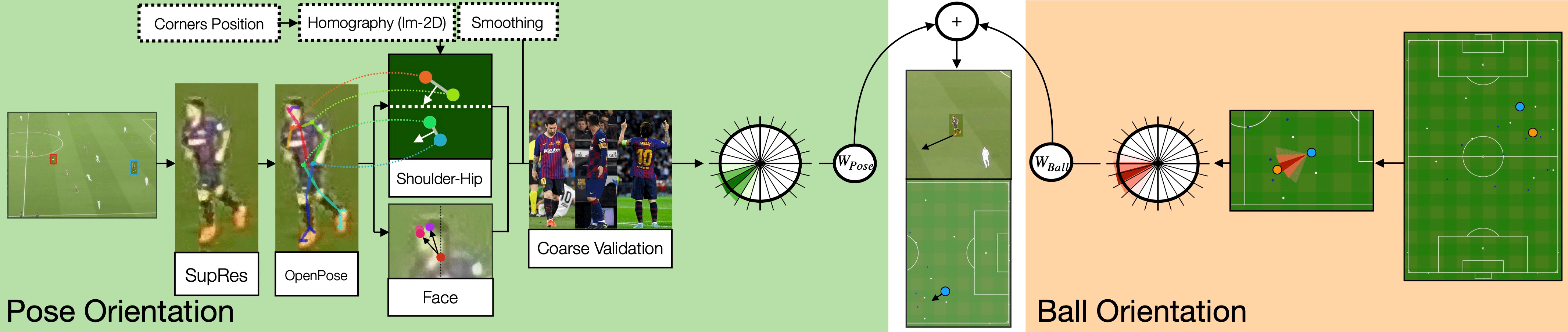}
  \caption{Proposed pipeline. On the one hand, pose orientation is found by combining a super-resolution network, OpenPose and 3D vision techniques (plus a coarse validation); on the other hand, ball orientation is also computed.}
  \label{fig:Pipeline}
\end{figure*}
In this paper, the orientation of a player's body is defined as the rotation of the player's upper-torso about the vertical axis, which is assumed to coincide with the angle of the 2D field projection of a 3D normal vector placed in the center of their upper-torso, involving both shoulders and hip parts. The overall pipeline of the presented method is displayed in Fig. \ref{fig:Pipeline}. This Section provides a detailed explanation of two different kinds of orientation estimation from which the method benefits: (1) pose data and (2) ball position. The output of all these individual estimations produces both a numerical orientation result and a confidence value. Orientation is measured in degrees and discretized into 24 probability bins using the reference system displayed in Fig. \ref{fig:References}(a). While the orientation value indicates the bin with higher probability, the confidence value is used as a prior to quantify, in an inversely proportional way, how many other neighboring bins have non-zero probability. This paper proposes an algorithm that outputs a probability density function (pdf) of the estimated orientation, thus containing both an estimated angle (maximum of pdf) and its confidence (inverse of the pdf support). A posteriori, a contextual weighting is performed to finally output the orientation of each player. 


\subsection{Pose Orientation}
Estimating orientation from pose data is a key ingredient of our method, and uses pretrained models and 3D vision techniques in order to obtain a first orientation estimation of each player. Given temporally-smoothed bounding boxes of players, a combination of super-resolution and pose detection techniques is applied to find the pose of every player. Both the left-right shoulders and the left-right parts of the hip will be considered as the main upper-torso parts. By projecting these parts in a 2D space, the normal vector between these points can be extracted. 

\textbf{Pose Detection}: having the bounding boxes for all visible players in each frame, the OpenPose library \cite{openpose} can be used to extract the pose of every single individual (we refer to \cite{ramakrishna2014pose, wei2016convolutional, cao2017realtime} for details of pose models). Given a soccer frame, the output of the pose estimator is a 25$\times$3 vector for each player, with the position (in image coordinates) of 25 keypoints, which belong to the main biometric human-body parts, together with a confidence score.
However, detecting the pose of players in sports scenarios is always challenging given the frequent occlusions and fast movements that lead to motion blur. Moreover, the average resolution of bounding boxes around players in Full-HD frames is around 15$\times$50 pixels. Hence, small image crops are not always properly processed by OpenPose, resulting in a null set of landmarks. For this reason, a super-resolution network is used to preprocess bounding boxes and enhance the image quality instead of a simpler interpolation technique. More concretely, the applied model is a Residual Dense Network (RDN) \cite{cardinale2018isr,zhang2018residual}. 

\begin{figure}
    \centering
    \includegraphics[width=0.4\textwidth]{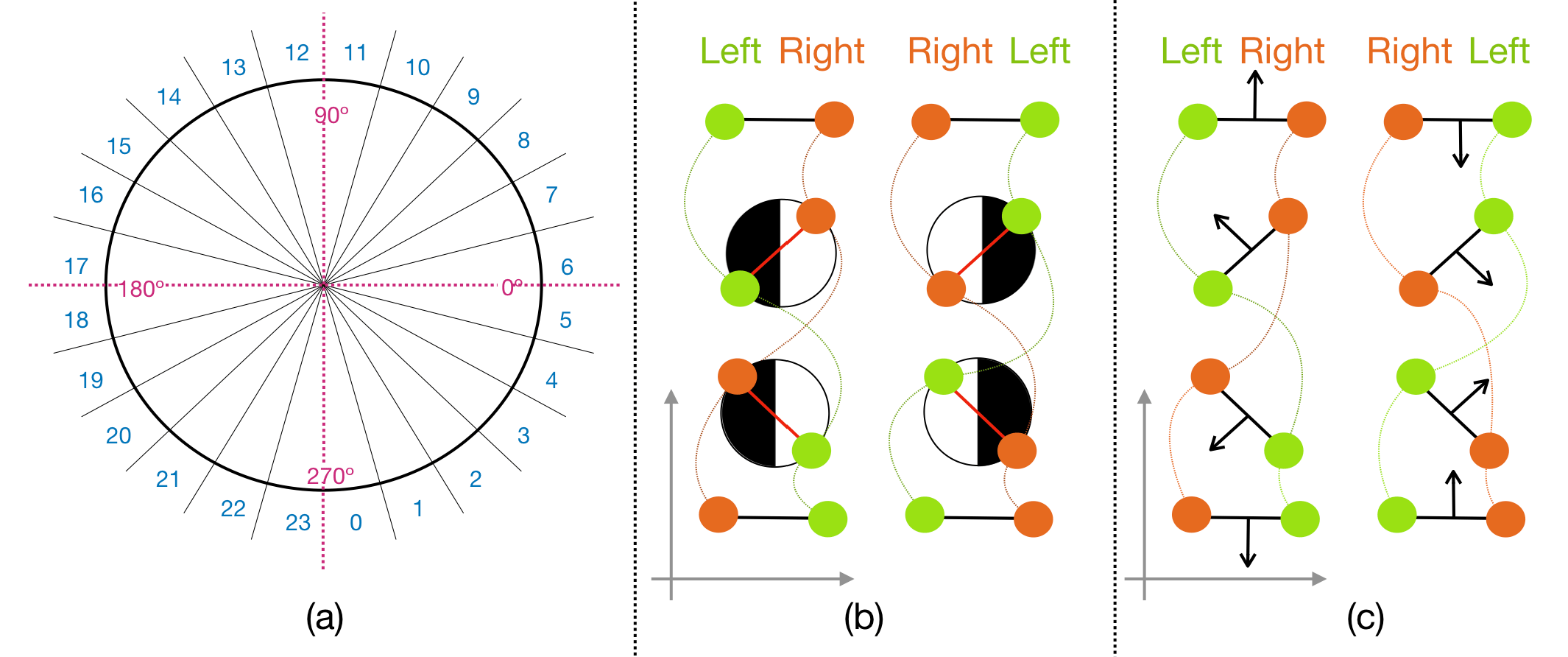}
    \caption{(a) Orientation discretization in 24 bins (blue - bin number). (b) Different 2D combinations of left-right mapped parts; (c) same combinations with normal vectors.}
    \label{fig:References}
\end{figure}

\textbf{Angle Estimation}: once the pose is extracted for each player, the coordinates (and confidence) associated to the upper-torso parts are stored to estimate the pose orientation. Given the four field corners' coordinates in the image plane,  
a homography is computed with the DLT algorithm \cite{hartley2003multiple} after establishing the four field-corners correspondences between the frame and a 2D field given by a template image of it. Other homography estimation strategies can be used (e.g., \cite{chen2019sports}). From the output of OpenPose, the coordinates of the main upper-torso parts are found in the image domain; by mapping the left-right pair (either shoulders or hips) in the 2D field, a first insight of the player orientation is obtained, as seen in Fig. \ref{fig:References}(b). Basically, the player can be inclined towards the right (0-90º, 270-360º, bins 0-11) or the left (90-270º, bins 12-23) side of the field. From now on, this first binary estimation, which indicates if the orientation belongs to the first or second half of the orientation histogram, will be called \textit{LR-side} parameter.

\begin{figure}
    \centering
    \includegraphics[width=0.45\textwidth]{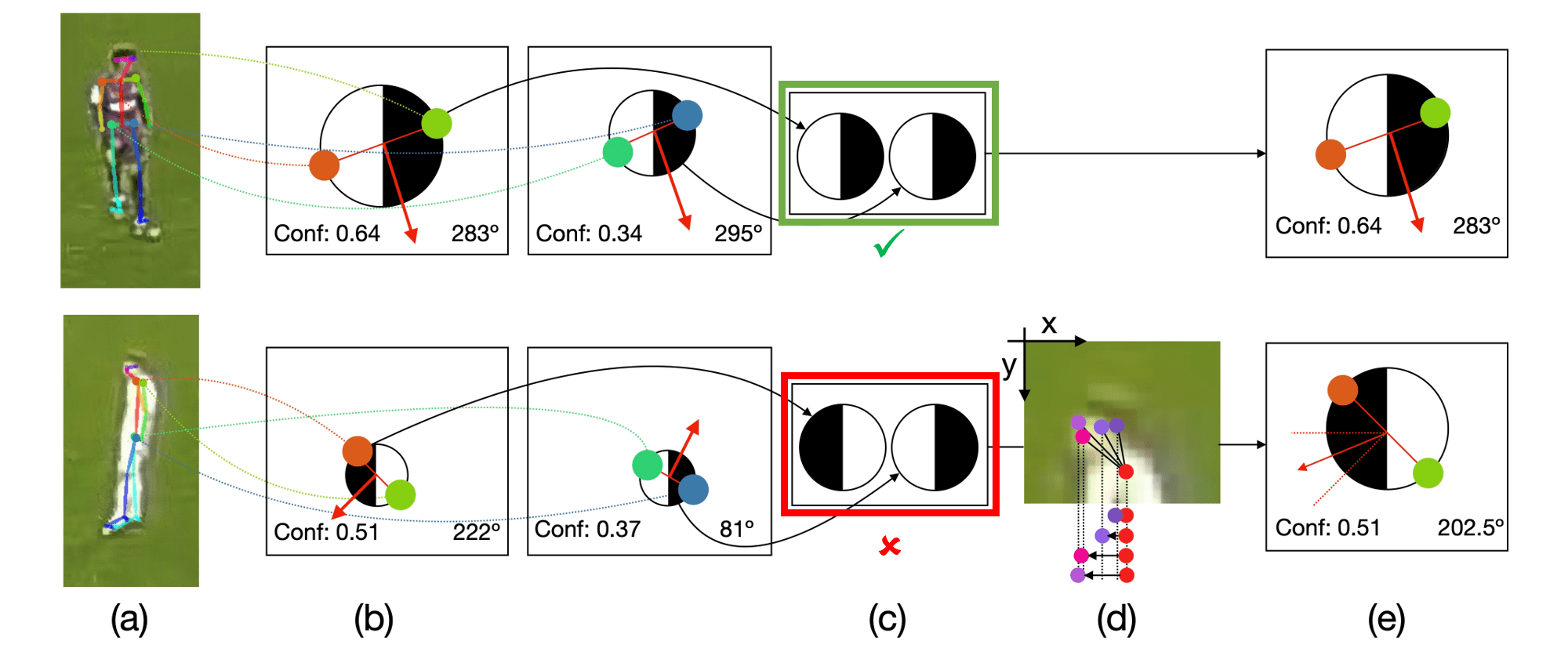}
    \caption{Pose orientation estimation: (a) OpenPose output, (b) mapped 2D coordinates. (c) Side check, (d) face direction double-check (e) a final estimation.}
    \label{fig:PoseOr}
\end{figure}

Fig. \ref{fig:PoseOr} shows in more detail how pose orientation is estimated: first, left-right shoulders and hips are mapped via the estimated homography into the 2D space; then, \textit{LR-side} booleans (${LR}_{Sh}$, ${LR}_{Hi}$), angles ($\alpha_{Sh}$, $\alpha_{Hi}$) and confidences ($C_{Sh}$, $C_{Hi}$) are obtained, where the suffixes $Sh$ and $Hi$ stand for shoulders and hips, respectively. The associated confidences are the product of OpenPose's individual shoulder and hips confidences respectively. However, OpenPose might fail detecting either the left or right hip parts; in these cases, the middle hip position is used as a substitute for the missing part. Then: \\
1. If ${LR}_{Sh}$ and ${LR}_{Hi}$ agree: If $C_{Sh} > C_{Hi}$, $\alpha_{Sh}$ is considered as the pose orientation estimation and $C_{Sh}$ its confidence. If not, $\alpha_{Hi}$ and $C_{Hi}$ are selected. \\
2. Otherwise, if $|C_{Sh} - C_{Hi}|$ is smaller than a threshold (\textit{i.e.} 0.4 in our results), the player's face direction is checked. In the image domain, the difference among the X positions of all face parts and the player's neck is computed. If most of the parts move towards the origin of the X axis (Fig. \ref{fig:PoseOr}(c)), the player's \textit{\textit{LR-side}} will be left; otherwise, the player's \textit{\textit{LR-side}} will be right.

Then, given the final pose orientation estimation $\alpha_{P}$ and its related confidence $C_{P}$, a Gaussian probability distribution is located around it, with effective support size
$
{
    N_{P} = \text{max}\left( \left\lfloor N_{bins} \left(\frac{1-C_{P}}{2} \right)\right\rfloor,1\right),
    }
$
centered at
$$ 
{\scriptstyle
\textrm{or}_{P} = \begin{cases}
\left\lfloor \frac{\alpha_{P}}{360/N_{bins}} + \frac{N_{bins}}{4} \right\rfloor & \; \textrm{if } \frac{\alpha_{P}}{360/N_{bins}} < 18
\\[10pt]
\left\lfloor \frac{\alpha_{P}}{360/N_{bins}} + \frac{N_{bins}}{4} \right\rfloor - N_{bins} & \; \textrm{if } \frac{\alpha_{P}}{360/N_{bins}} \geqslant 18
\end{cases}
}
$$
where the second element of the sum is an offset that compensates the bin order (Fig. \ref{fig:References}(a)). The output vector of this orientation estimation will be denoted as $H_{P}$.

\begin{figure}
    \centering
    \includegraphics[width=0.35\textwidth]{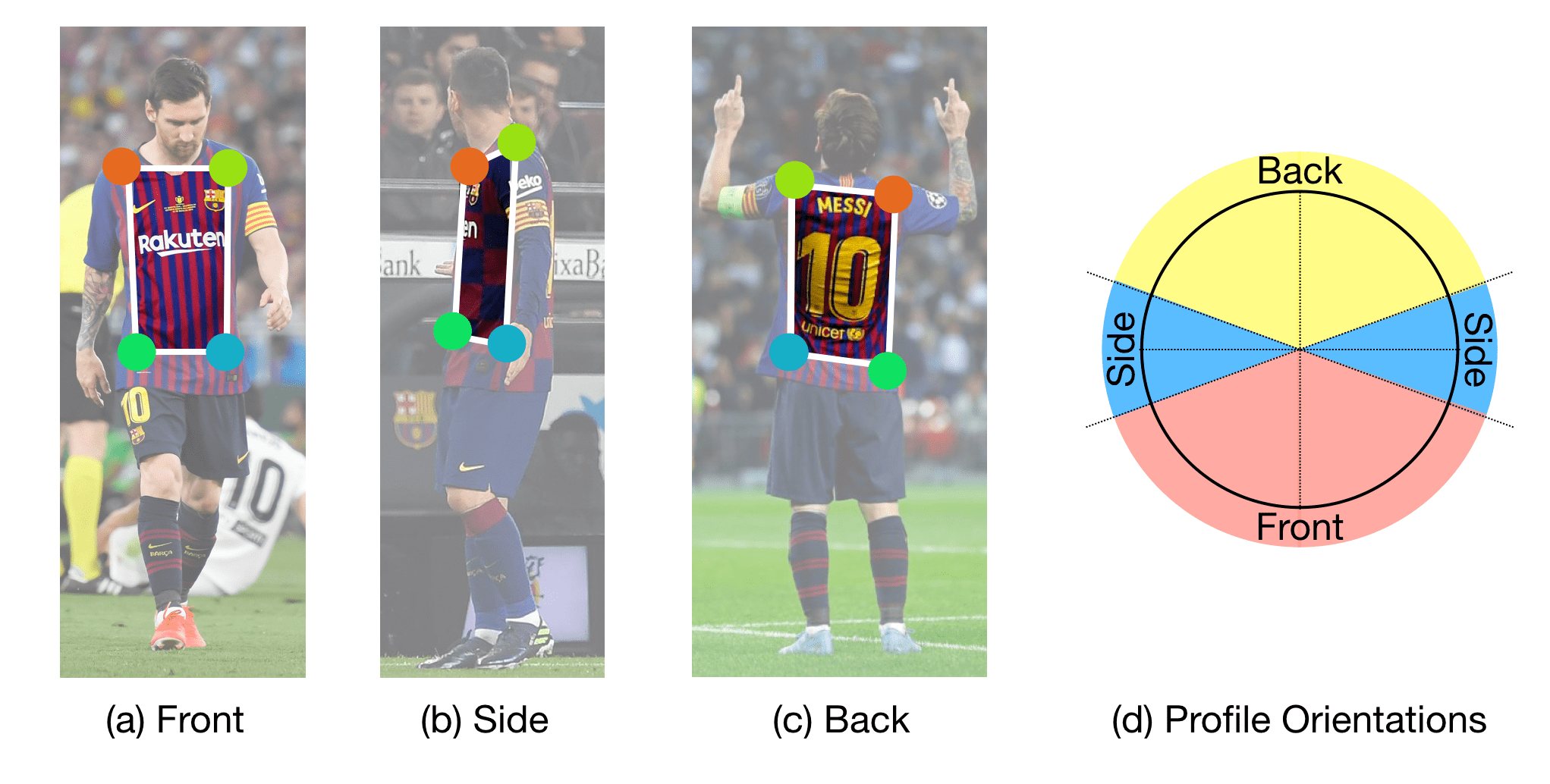}
    \caption{(a) front-, (b) side-, and (c) back-oriented players, (d) corresponding potential pose orientation.}
    \label{fig:CoraseOr}
\end{figure}

\textbf{Coarse Orientation Validation}: despite the notable performance of Open Pose, image quality problems (e.g. blurry or really small players) are challenging scenarios where estimated players’ pose might be flipped 180º: this is, the right-left shoulders (or hip parts) of the corresponding player are swapped. An inaccurate detection of the player pose results in huge errors while estimating the pose angle, as the actual normal vector is the opposite of the predicted one, thus introducing errors that might oscillate between 120º and 180º. In order to double-check the pose orientation estimation and to ensure that the upper-torso normal vector is computed in the correct direction, a Support Vector Machine model has been trained to classify three types of coarse orientations: front-, side- and back-oriented players (see Fig. \ref{fig:CoraseOr}).
Two characteristics are concatenated in the feature vector: color features in the Hue-Saturation-Value color space (histogram of 36-18-18 bins in the respective channel) and geometrical properties (pixel-wise distances between the 4 upper-torso coordinates). Having the position of the upper-torso parts, obtained from pose keypoints, the above-mentioned features are only computed inside the defined trapezoid, hence discarding misleading features such as the color of the field. 

\begin{figure*}
\centering
  \includegraphics[width=0.85\textwidth]{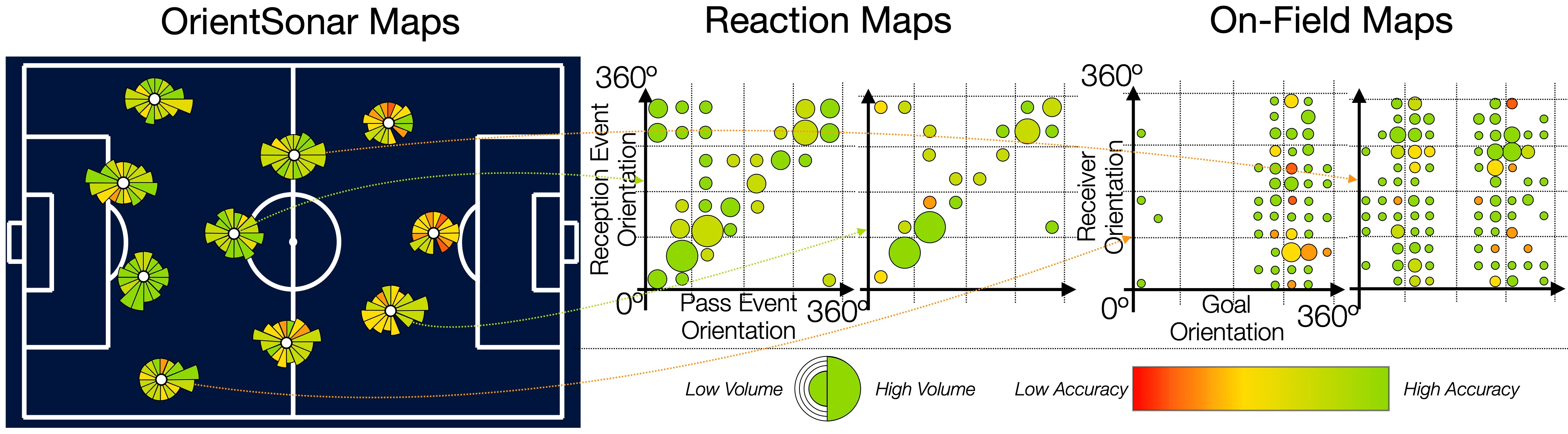}
  \caption{(left) team-level OrientSonar (\textit{Reception Events}), (mid.) individual reaction map, and (right) individual on-field map.}
  \label{fig:Maps}
\end{figure*}

\subsection{Ball Orientation}
The other performed estimation is related to the position of the ball. Logically, players close to the ball tend to be strongly oriented towards it, while players placed far away may not have to be duly oriented accordingly. Hence, having all pairwise distances and the corresponding angles, the orientation of players with respect to the ball can be estimated. Then, for a given player at $(P_{x},P_{y})$, in a moment where the ball is at $(B_{x},B_{y})$ generating an angle of $\beta$ degrees, the effective support size of the related pdf is: 
$
{
N_{B} = \frac{N_{bins}}{4} \left\lfloor 1 - \frac{MD-\sqrt{(P_{x}^2-B_{x}^2)+(P_{y}^2-B_{y}^2)}}{MD} \right\rfloor + \frac{N_{bins}}{8}},
$
where $MD$ is a maximum distance that regularizes how far a player can be from the ball without being influenced by it. Then, the central bin with the highest weight is: 
$$ 
{\scriptstyle
\textrm{or}_{B} = \begin{cases}
\left \lfloor \frac{\beta}{360/N_{bins}} + \frac{N_{bins}}{4} \right  \rfloor  & \; \textrm{if } \left \lfloor \frac{\beta}{360/N_{bins}} \right \rfloor < 18
\\[10pt]
\left \lfloor \frac{\beta}{360/N_{bins}} + \frac{N_{bins}}{4} \right \rfloor - N_{bins}  & \; \textrm{if } \left \lfloor \frac{\beta}{360/N_{bins}} \right \rfloor \geqslant 18
\end{cases}
}
$$
Once again, the outcome of this estimation is a discretized probability vector, called from now on $H_{B}$. 
\subsection{Contextual Merging}
Once both histograms are obtained, a simple weighting is performed between them, thus merging pose and ball orientations. In particular: 
$
H_{\text{TOT}} = w  H_{P} + (1-w) H_{B}, 
$
with $w \in [0,1]$.
 The orientation $\theta$ of each player is the central value of the bin $H_\text{TOT}$ with higher weight.   

\section{Results}
The dataset provided by F.C. Barcelona included video footage (25 fps) of several games from La Liga, tracking data in both frame and field domains, corners positions and contextual information. Moreover, XYZ orientation data were gathered from youth games using EPTS devices \cite{fifaEpts} for quatitative assessment. 

Bearing in mind that OpenPose detected upper-torso parts in \textbf{89.69\%} of the given image crops: \\ 
\textbf{Coarse orientation validation}: 14,000 players were manually labelled (front, back or side); by randomly splitting it into train and test (80-20), 85.91\% accuracy was obtained. \\ 
\textbf{\textit{LR-side}:} this metric shows the accuracy of the \textit{LR-side} parameter, which indicates if a player is facing the left or the right side of the field. Considering a sequence of duration $T$ and being $i_t$ an individual player in a total of $NP_{t}$ players in frame $t$, pose orientation $\alpha_{i_t}$, and the corresponding ground-truth orientation $\omega_{i_t}$, this metric can be computed as: 
$$
{
\text{LR}_\text{acc} = \frac{\sum_{t=0}^{T}\sum_{i_t=0}^{NP_{t}}LRV_{i_t}}{\sum_{t=0}^{T}NP_{t}} }
$$
where:
$$ 
{\scriptstyle
LRV_{i_t} = \begin{cases}
1 & \; \textrm{if} \, |\alpha_{i_t}-\omega_{i_t}| < |\alpha_{i_t}+180-\omega_{i_t}|\\
0 & \text{otherwise}
\end{cases}}
$$
\textit{\textit{LR-side}} performance reached \textbf{96.57\%} accuracy. 


\textbf{Parameter Adjustment}: by testing all possible weight combinations (in 0.05 intervals), results in Table \ref{tab:tabRes} indicate the error margin of different tests, showing the performance of each individual orientation estimation and their best mixture. As it can be observed, ball orientation produces the less accurate predictions; pose orientation outperforms this prediction by a notable margin. These individual results prove that pose orientation needs to be heavily weighted while merging both estimations: by setting $w$ to 0.7, the mean absolute angle error (MEAE) is reduced to \textbf{29.78º} and the median absolute angle error (MDAE) to \textbf{27.66º}. 

\begin{table}[]
\centering
\scalebox{0.8}{
\begin{tabular}{|c|c|c|c|}
\hline
\textit{$w$}  & \textit{$(1-w)$}  & \textit{MEAE}  & \textit{MDAE}  \\ \hline
\textit{0}   & \textit{1}   & 35.33          & 31.59          \\ \hline
\textit{1}   & \textit{0}   & 29.98          & 27.75          \\ \hline
\textit{0.3} & \textit{0.7} & 33.77          & 29.87          \\ \hline
\textit{0.7} & \textit{0.3} & \textbf{29.78} & \textbf{27.66} \\ \hline
\end{tabular}
}
\caption{MEAE and MDAE given different weights.}
\label{tab:tabRes}
\end{table}
\section{Practical Applications}
In this Section, the visualization of raw orientation data in passing events is analyzed (12 games and 6836 passes were included in the dataset). In particular, the orientation of the potential receiver of a pass is computed (a) right when the passer kicks the ball (\textit{Pass Event}), and (b) at the exact moment where the player receives (\textit{Reception Event}). As seen in Fig.~\ref{fig:Maps}, three different types of maps can be extracted: \\
\textbf{OrientSonars} integrate player orientation and show how players are oriented during pass events. In this display, the following size-color codification is adopted: the radius length of each portion in the map quantifies the volume of passes at a particular orientation, while the color displays their associated accuracy. \textbf{Reaction Maps} show how players are moving during the pass, by comparing the orientation at the beginning and at the end of the event; once again, dot area expresses the volume. If a player keeps his/her orientation, the resulting map will just have dots in a diagonal line; otherwise, off-diagonal dots appear in the graph. 
\textbf{On-Field Maps} merge and compare the pure body orientation of players with their relative orientation with respect to the offensive goal. All these maps can be extracted at a player- or team-level, and custom filters can be created in order to introduce context such as game phases (build-up, progression and finalization); moreover, accuracy might not be the best metric to be used, so maps can be color-codified according to other techniques like Expected Posession Value \cite{fernandez2019decomposing}.   

\section{Conclusions}
In this article, a novel technique to compute soccer players' orientation from a video has been presented. The method combines two different orientation estimators: pose and ball. While pose orientation is obtained by projecting OpenPose output on a 2D space and computing the normal vector to the projected torso, ball estimation calculates the orientation of all players with respect to the ball.  Results have been tested and validated with professional soccer matches: 96.6\% accuracy is obtained in left-right side orientations, and a median absolute error of 27.66º is achieved. As future work, besides improving the MEAE, which could be done by training an end-to-end deep learning model, the generalization to other sports and scenarios will be studied. Furthermore, other applications using pose data could be tested (\textit{i.e.} team/individual action recognition). Finally, new types of visualizations could be designed together with complex game phases by including more information in the dataset.  





\appendix

\section{Orientation Maps}
Since the original document is publication of 4 pages, this appendix tries to complement the presented applications in the main paper: orientation maps. Please refer to the previous paper Sections for a detailed technical methodology of the proposed model.\\ 
As it is introduced in the article, obtaining  orientation metrics may help coaches to boost the performance of a team by designing optimal tactics according to players’ strengths and weaknesses. However, orientation is not an easy statistic to handle when dealing with raw data, so eventing filtering and visual displays might be the tools that coaches need to have this feature under control. \\
In this section, the effect of body orientation is analysed in soccer passing events. In particular, three types of cases are considered:
\begin{enumerate}
    \item \textbf{Orientation of the receiver in a \textit{Pass Event}}: this value quantifies the orientation in the field of a potential receiver of a pass right at the moment when the passer kicks the ball.
    \item \textbf{Orientation of the receiver in a \textit{Reception Event}}: this value quantifies the orientation in the field of a player who is receiving the ball at that precise moment.
    \item \textbf{Orientation of the passer in a \textit{Pass Event}}: this value quantifies the orientation of the player kicking the ball when performing a pass.
\end{enumerate}

Moreover, the following performance statistics are used in order to evaluate the impact of body orientation in the observed passes:
\begin{enumerate}
    \item \textbf{Pass success/accuracy}, which indicates if the pass was successful or not; this is, if the potential receiver has actually received the ball. This metric can be used to get an overall picture of orientation, but there might be a lack of context: an easy pass between two defenders is valued the same way as a difficult assist that ends up in a goal. Besides, a failed pass might happen due to multiple circumstances, such as a bad pass, a bad reception, or a remarkable performance of a defender.
    \item \textbf{Added Expected Possession Value (EPV)}, a recent state-of-the-art method introduced by Fernandez \textit{et al.} \cite{fernandez2019decomposing} that quantifies the contribution of each action by modelling the conditional probability of scoring/receiving a goal at a given time and a given scenario. EPV is computed both at the \textit{Pass Event} and right after the \textit{Reception Event}; the difference between these two values will indicate the added contribution of the receiving player and exemplifies what happens after receiving the ball. For instance, a player might receive the ball appropriately but he/she might lose it due to a disadvantageous orientation, resulting in an EPV drop.
\end{enumerate}

In order to introduce context in the mentioned visualizations, different phases of the offensive plays are evaluated individually as well. Bearing in mind that in a soccer lineup there are mainly 3 rows of horizontally distributed players, their orientation can drastically change depending on the context: if an almost-static defender is carrying the ball, strikers will not be strictly oriented towards it, but if a midfielder is generating a play in the offensive court, forwards will be highly influenced by his position. Moreover, the role of all defensive players also take a crucial role in the decision-making process. By clustering the 2D coordinates of the players in the field, the ball position can be found in three states or phases:

\begin{enumerate}
    \item \textbf{Build-up} phase: the ball is located before the first row of defensive players. 
    \item \textbf{Middle} phase: the ball is located between the first and the second row of defensive players. 
    \item \textbf{Progression} phase: the ball is located after the second row of defensive players.
\end{enumerate}

\begin{figure}[H]
    \centering
    \includegraphics[width=0.4\textwidth]{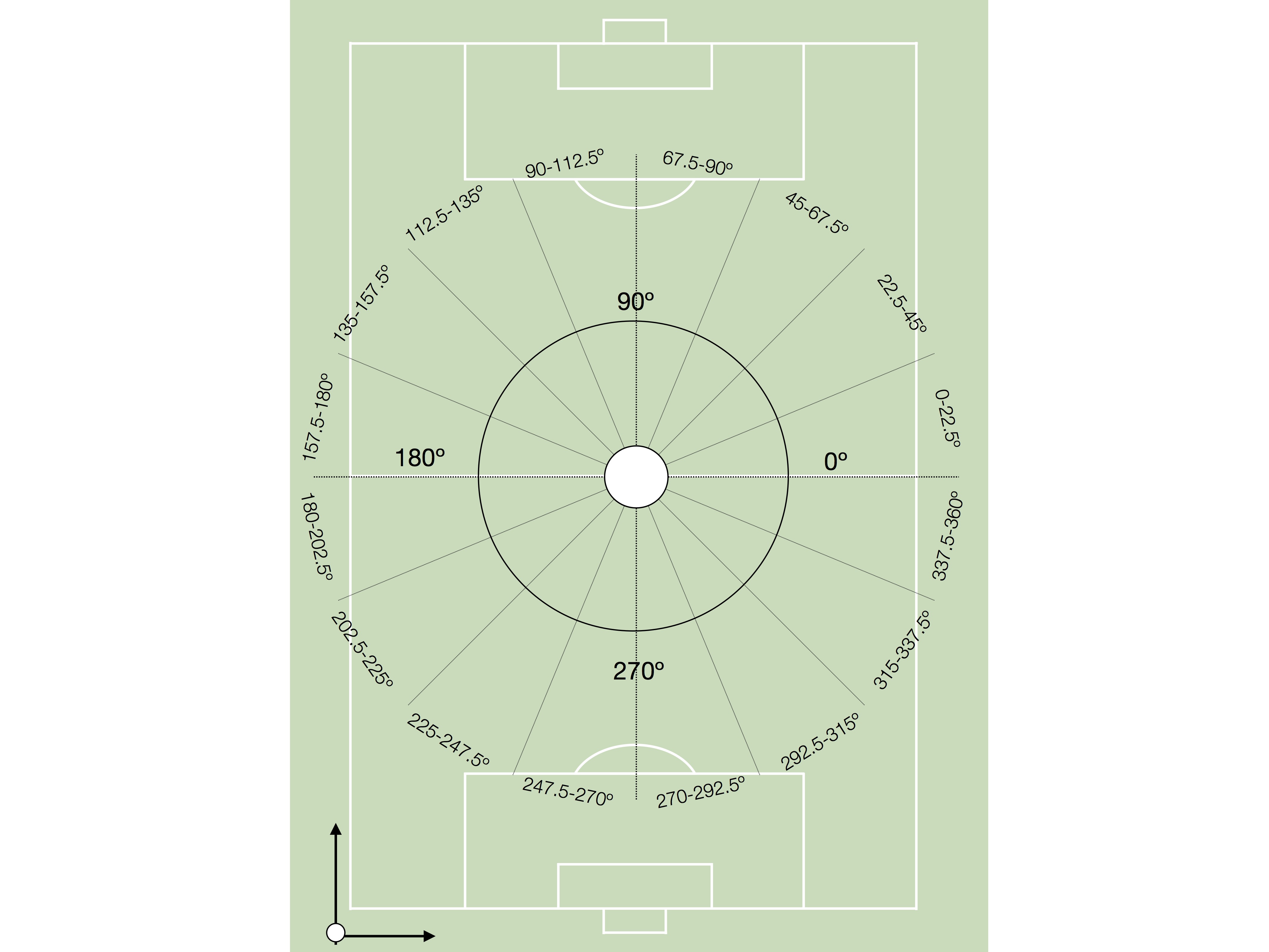}
    \caption{Orientation references in the soccer field.}
    \label{fig:OrFirst}
\end{figure}

By filtering data from F.C. Barcelona games during the 2019-2020 season, 7500 event passes have been gathered among 12 different players; orientation and performance metrics have been computed for each one in order to create helpful visualizations. For the rest of this Section, orientations have been estimated using the reference system shown in Figure \ref{fig:OrFirst}.

\section{OrientSonars}
PassSonars have recently gained a lot of popularity in soccer analytics; this kind of map is used to display the passing frequency and accuracy of players in different directions inside the field, just by taking 2D information from these. In this article, OrientSonars are proposed, which integrate player orientation and show how players are oriented during pass events. In this display, the following size-color codification is adopted: the radius length of each portion in the map quantifies the volume of passes at a particular orientation, while the color displays their associated accuracy. OrientSonars can be performed at two levels:\\ 

\begin{figure}[H]
    \centering
    \includegraphics[width=0.4\textwidth]{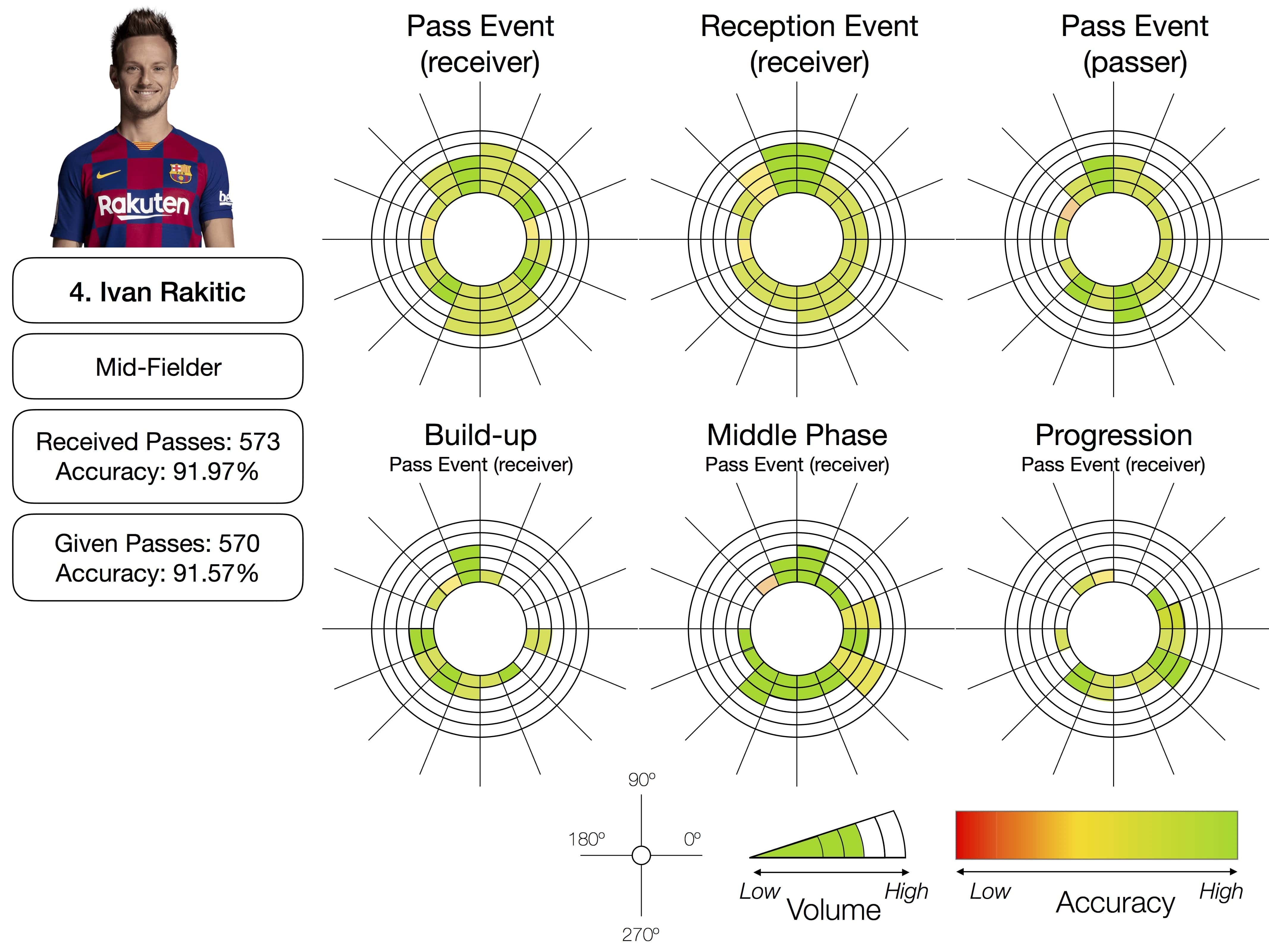}
    \caption{OrientSonar of Ivan Rakitic, showing his performance in pass events (both passing and receiving the ball) and reception ones, as well as different offensive phases. Accuracy is expressed with pass accuracy and color encoding, while portion size indicates the passing volume.}
    \label{fig:OrSonarPlayer}
\end{figure}

\textbf{Individual level}: simple visual reports of each player can be built by combining different OrientSonars in the 3 above-mentioned possible events. These visualizations can be useful to spot specific details when scouting a particular player. An example is shown in Figure \ref{fig:OrSonarPlayer}, where the main orientation characteristics of Ivan Rakitic are shown. \\

In this specific example it can be seen that Rakitic, as a mid-fielder, has a strong duality receiving passes when oriented completely backwards (~270º) and upwards (~90º), as he has to receive passes from defenders (backwards) and organize the forwards at the same time. In particular, Rakitic excels in reception events when the orientation oscillates between 67.5 and 112.5 degrees, which matches the most natural reception orientation for right footed player. Moreover, game phases indicate that Rakitic is oriented towards defenders in the build-up phase (especially the left-side ones), but when the ball is carried towards the middle of the court, he is also oriented towards the offensive goal, thus potentially generating passes to forwards. \\

\begin{figure*}
\centering
\includegraphics[width=\textwidth]{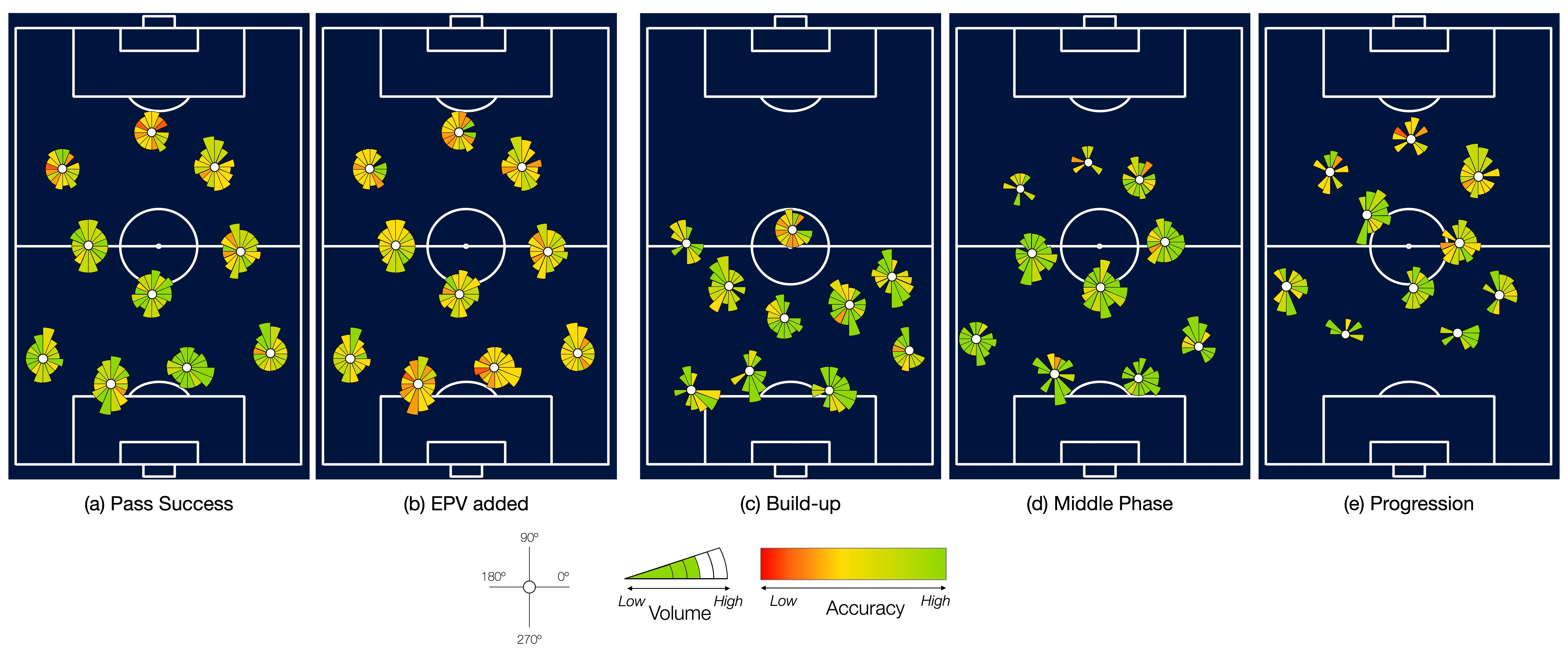}
\caption{OrientSonar of the whole team during \textit{Pass Events} as receivers, displayed with different accuracy metrics: (a) pass success metrics, and (b) added EPV. Moreover, (c) build-up, (d) middle-phase and (e) progression are analyzed individually with pass success metrics.}\label{fig:OrSonarTeam}
\end{figure*}

\textbf{Team level}: as individual performances might be biased towards specific team tactics, the whole picture of the corresponding lineup has to be evaluated as well. In this map, the individual OrientSonar of all players is placed at the average position of every single individual. An example can be seen in Figure \ref{fig:OrSonarTeam} (a,b), where accuracy and added-EPV are compared, and Figure \ref{fig:OrSonarTeam} (c,d,e), where different games phases can be distinguished. \\

Several conclusions can be drawn from these maps: from Figure \ref{fig:OrSonarTeam}(a,b), it can be proven that the pass success might not be the best accuracy metric to be used when comparing all kinds of players, mainly because defenders perform many non-risky passes among themselves, while forwards receive the ball in fewer situations (and often under the pressure of defenders) with higher risk and potential reward. For this reason, defenders in Figure  \ref{fig:OrSonarTeam}(a) have a lot of high-accuracy clusters, and forwardreceive less passes at a lower accuracy rate. This situation swaps when checking EPV: on the one hand, defenders add less real value to the play, and on the other hand, offensive players have some clusters with high contribution when they receive in advantageous situations (facing slightly upwards). Moreover, EPV peaks do not appear in random clusters: instead, a notable increment of EPV can be observed when specific couples of players interact. For instance, when the striker receives the ball from approximately the position of the right forward or vice versa, the team not only keeps the ball, but also creates potential goal opportunities. The same pattern is repeated with the center- and left-midfielder. Besides, orientation patterns may be useful to distinguish the dominant player side: left-sided players (i.e. left full back) tend to be oriented towards the middle of the field, so right-side clusters have higher volume of passes (and vice versa). \\

From Figure \ref{fig:OrSonarTeam} (c,d,e), the interaction of players is even more detailed according to the context: in the build-up phase, midfielders are orientated towards the defenders, waiting for the ball in order to generate offense. In the middle phase, the same midfielders have the higher relevance in terms of volume, distributing the ball in potentially advantageous situations; meanwhile, strikers look for open spaces, and rarely receive the ball backwards (except the right-forward in the given example). Finally, in the progression phase, two possible player roles can be distinguished: while some players are oriented towards regions with high risk but a notable potential reward, the rest occupy safe positions that allow them to move back to another medium phase if required without losing control of the ball, thus generating new offensive opportunities. \\

\section{Orientation Reaction Maps}
\begin{figure}
    \centering
    \includegraphics[width=0.4\textwidth]{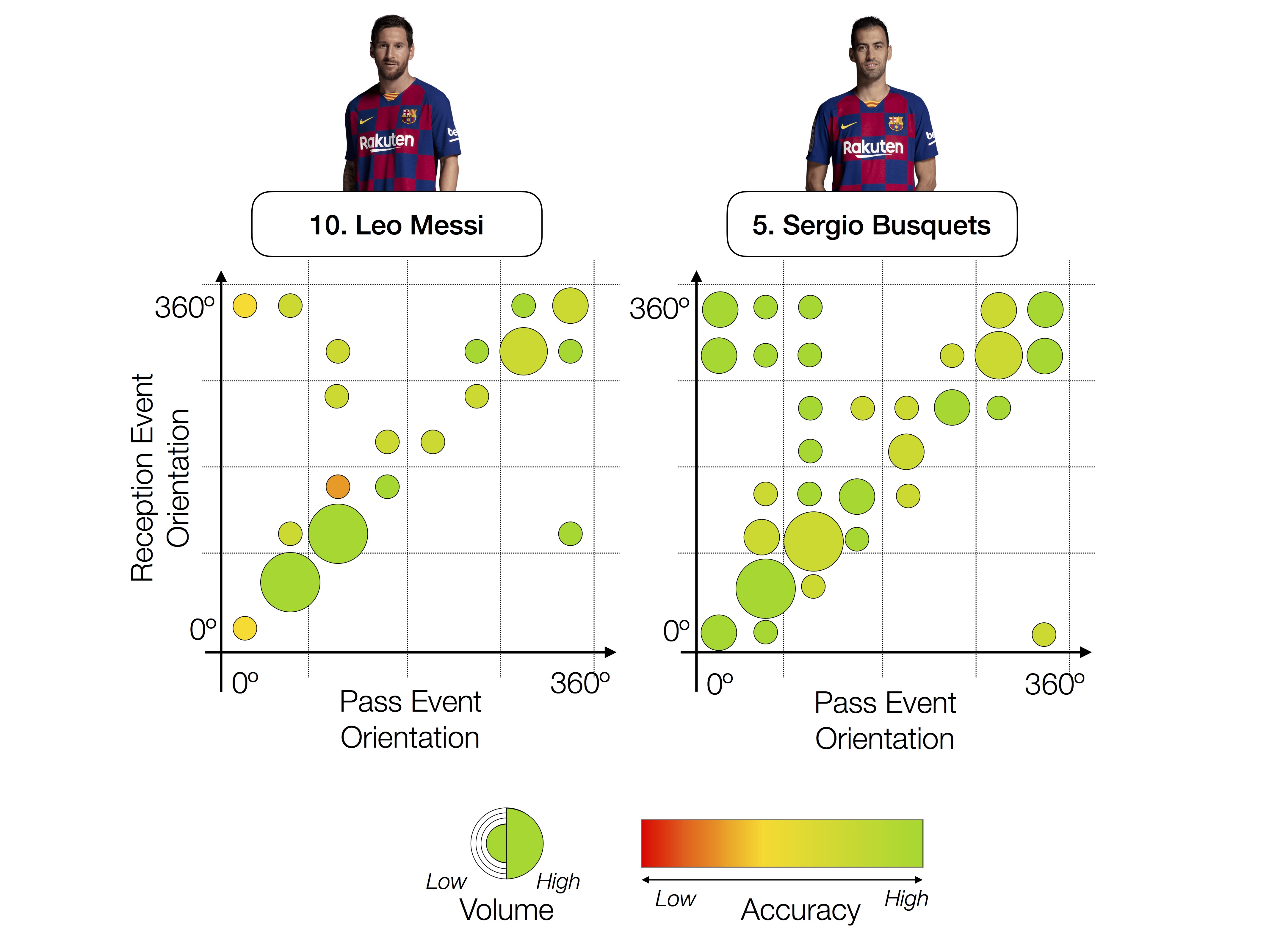}
    \caption{(left) Leo Messi – (right) Sergio Busquets reaction maps. The X axis represents the orientation of the player at the pass event, and the Y axis the one in the reception. Accuracy has been computed with expressed with pass success.}
    \label{fig:ReactionTime}
\end{figure}

Although there are many different types of soccer passes, the behavior of players during the ball displacement is crucial to the outcome of that specific play; defenders are always trying to anticipate, so offensive players must orient and move accordingly before getting tackled. Orientation reaction maps show how players are moving during the pass, by comparing the orientation at the beginning (X axis) and at the end (Y axis) of the event; once again, the color represents accuracy, and dot area expresses the volume. If a player keeps his orientation, the resulting map will just have dots in a diagonal line; on the contrary, if a player rotates while receiving, off-diagonal dots appear in the graph. Figure \ref{fig:ReactionTime} shows the orientation reaction maps of Messi (right forward) and Busquets (central midfielder). \\

Once again, the visual outcome differs for those players who occupy different positions. In the given example, Messi has a main diagonal line with some outliers, as he receives many passes from players who are in front of him (facing backwards) when he is running towards the goal (huge blobs in the 75-105º), as well as straight passes from midfielders when his is facing backwards (270-315º). Meanwhile, Busquets has more dots in his map, mainly because he receives passes from many different positions; besides, being close to the exact middle point of the field makes things even trickier, as he has defenders trying to tackle him from several positions, thus forcing him to move even more to find a safe spot. As it can be seen in the map, Busquets has a remarkable performance in every single orientation, especially in the right-side clusters.
In conclusion, there is not an optimal reaction map, and comparisons have to be performed by contextualizing the player position in the field, as it is difficult to establish similarities among players with different characteristics.

\section{On-Field Orientation Maps}
Despite being the goal the most important part in soccer games, all the previous displays were only based on the orientation of the player at given events. Hence, proposed on-field orientation maps merge information and compare the pure body orientation of players with their relative orientation with respect to the offensive goal. Given this scenario, the orientation with respect to the goal will be computed using the same reference system shown in Figure \ref{fig:OrFirst}. \\

\begin{figure}
    \centering
    \includegraphics[width=0.4\textwidth]{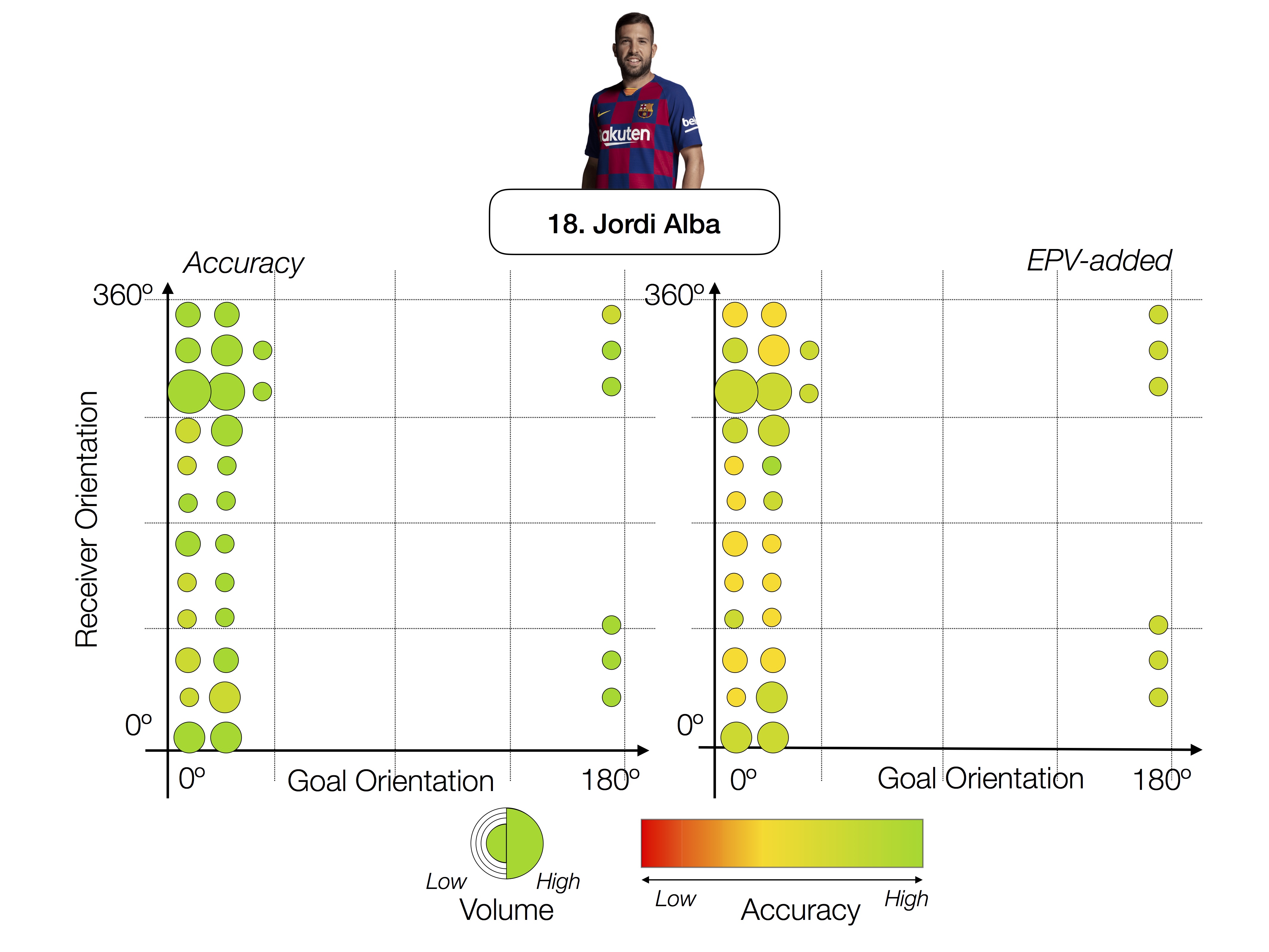}
    \caption{On-field orientation map of Alba as a receiver in \textit{Pass Events}, evaluated both with (left) pass accuracy and (right) EPV metrics.}
    \label{fig:OnFieldAlba}
\end{figure}

\begin{figure}
    \centering
    \includegraphics[width=0.4\textwidth]{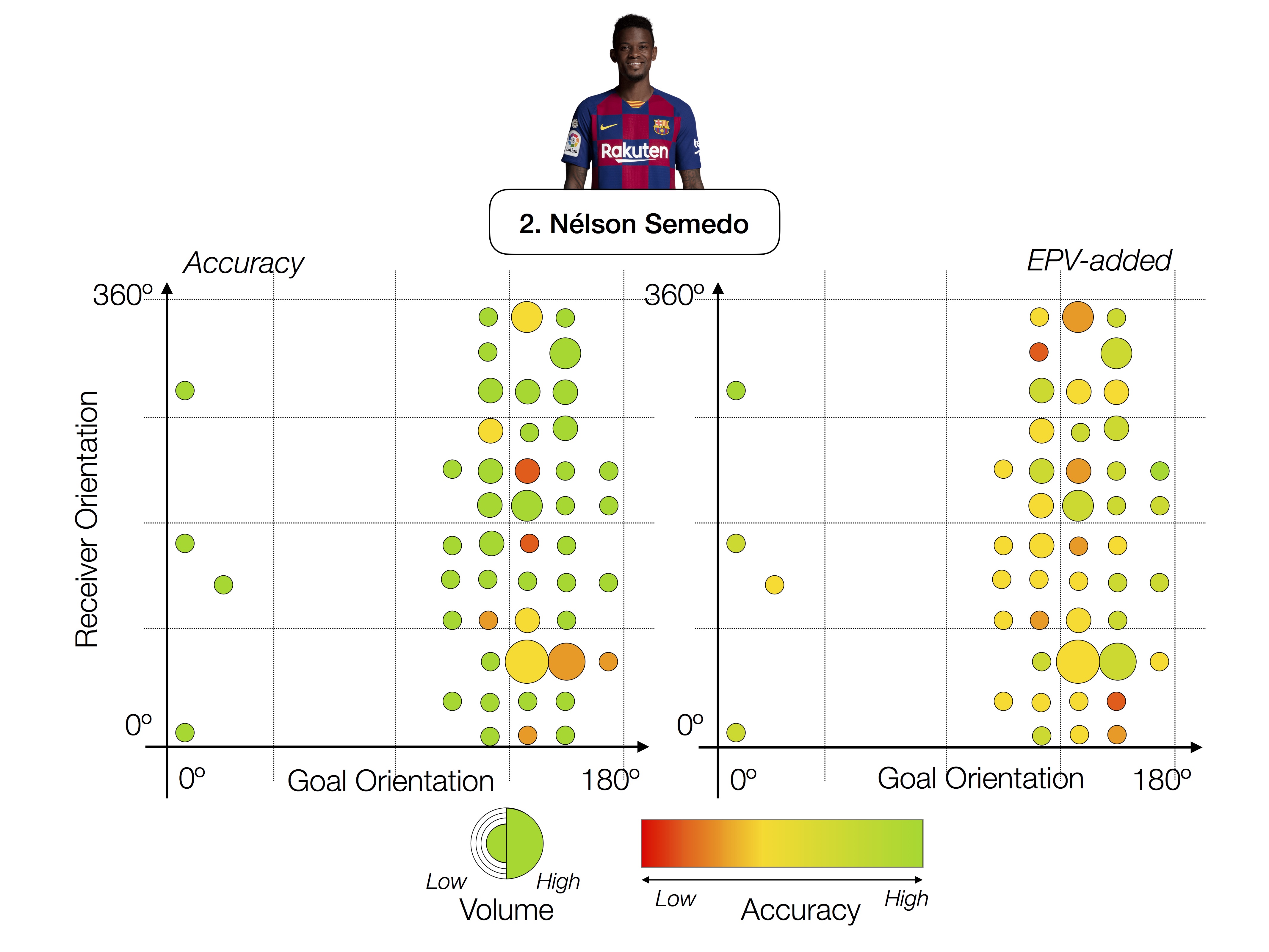}
    \caption{On-field orientation maps of Semedo as a receiver in \textit{Pass Events}, evaluated both with (left) pass accuracy and (right) EPV.}
    \label{fig:OnFieldSemedo}
\end{figure}

\begin{figure}
    \centering
    \includegraphics[width=0.4\textwidth]{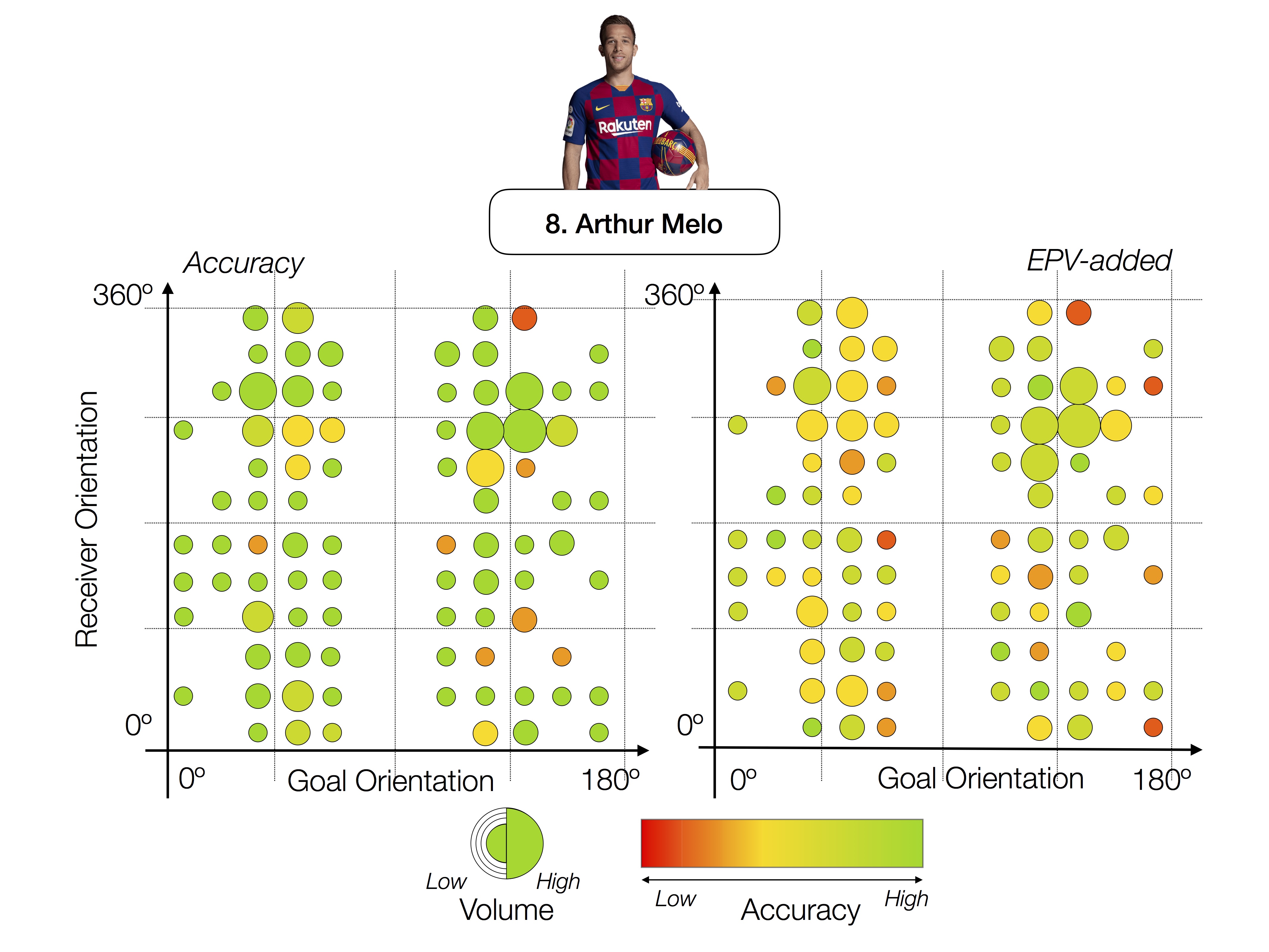}
    \caption{On-field orientation maps of Arthur as a receiver in \textit{Pass Events}, evaluated both with (left) pass accuracy and (right) EPV.}
    \label{fig:OnFieldArthur}
\end{figure}

Visual maps can be extracted at a player-level, as seen in Figures \ref{fig:OnFieldAlba}, \ref{fig:OnFieldSemedo}, \ref{fig:OnFieldArthur}, where both left-right full backs (Alba – Semedo) are compared and another individual performance of a midfielder (Arthur) is shown as receivers of \textit{Pass Events}. In these visualizations, the X axis represents the orientation with respect to the offensive goal (being 0-90 the left side and 90-180 the right side), and the Y axis represents the orientation of the player.

In this type of map, it is even more distinguishable how are players clustered depending on their position. Despite the difference in spatial performance, the visualizations of Alba and Semedo show almost symmetric results for left- and right-sided players. While Alba is completely restricted to the left side of the court (0-45º in goal orientation), Semedo tends to deviate his orientation more towards the middle part of the court, which results in regions with an EPV drop. This particular scenario shows one of the main differences between experienced players and the rest: in this case, bearing in mind that Jordi Alba has been on the team for 7 seasons in a row, it is reasonable to conclude that he already found a comfort zone in court, where he manages to fit all his skills without the need of taking unnecessary risks. \\
Apart from the fullback comparison, the plot of Arthur shows that this type of midfielder operates on both the central-sides of the court at more or less the same frequency; although orientation performance could seemingly be the same when checking pass accuracy, EPV can help detecting complex patterns. In the given example, Arthur adds higher EPV contributions when he is placed in the right-side of the court, specially when receiving in a backwards orientation (most likely from a defender); nevertheless, this type of conclusion has to be again contextualized with different prior information (i.e., a player that just started playing in a new spot for the first time in the season and needs some adaptation). Filtering again by game phases could enrich this type of map; however, data from at least a whole season would be required in order to display meaningful and confident results.


\begin{thebibliography}{10}

\bibitem{openpose}
{OpenPose}.
\newblock \url{https://github.com/CMU-Perceptual-Computing-Lab/openpose}.

\bibitem{cao2017realtime}
Z.~Cao, T.~Simon, S.-E. Wei, and Y.~Sheikh.
\newblock Realtime multi-person 2d pose estimation using part affinity fields.
\newblock In {\em IEEE Conference on Computer Vision and Pattern Recognition},
  pages 7291--7299, 2017.

\bibitem{cardinale2018isr}
F.~Cardinale.
\newblock Isr.
\newblock \url{https://github.com/idealo/image-super-resolution}, 2018.

\bibitem{chen2019sports}
J.~Chen and J.~J. Little.
\newblock Sports camera calibration via synthetic data.
\newblock In {\em Proceedings of the IEEE Conference on Computer Vision and
  Pattern Recognition Workshops}, pages 0--0, 2019.

\bibitem{fastovets2013athlete}
M.~Fastovets, J.-Y. Guillemaut, and A.~Hilton.
\newblock Athlete pose estimation from monocular tv sports footage.
\newblock In {\em IEEE Conference on Computer Vision and Pattern Recognition
  Workshops}, pages 1048--1054, 2013.

\bibitem{felsen2017will}
P.~Felsen, P.~Agrawal, and J.~Malik.
\newblock What will happen next? forecasting player moves in sports videos.
\newblock In {\em IEEE ICCV}, pages 3342--3351, 2017.

\bibitem{felsen2017body}
P.~Felsen and P.~Lucey.
\newblock Body shots: Analyzing shooting styles in the {NBA} using body pose.
\newblock In {\em MIT Sloan, Sports Analytics Conference}, 2017.

\bibitem{fernandez2019decomposing}
J.~Fern{\'a}ndez, L.~Bornn, and D.~Cervone.
\newblock Decomposing the immeasurable sport: A deep learning expected
  possession value framework for soccer.
\newblock In {\em 13th MIT Sloan Sports Analytics Conference}, 2019.

\bibitem{fifaEpts}
FIFA.
\newblock {EPTS Electronic Performance and Tracking Systems}.
\newblock \url{https://football-technology.fifa.com/en/media-tiles/epts/}.

\bibitem{fischer2018rt}
T.~Fischer, H.~Jin~Chang, and Y.~Demiris.
\newblock Rt-gene: Real-time eye gaze estimation in natural environments.
\newblock In {\em European Conference on Computer Vision (ECCV)}, pages
  334--352, 2018.

\bibitem{gao2019graph}
J.~Gao, T.~Zhang, and C.~Xu.
\newblock Graph convolutional tracking.
\newblock In {\em IEEE Conference on Computer Vision and Pattern Recognition},
  pages 4649--4659, 2019.

\bibitem{girdhar2018detect}
R.~Girdhar, G.~Gkioxari, L.~Torresani, M.~Paluri, and D.~Tran.
\newblock Detect-and-track: Efficient pose estimation in videos.
\newblock In {\em IEEE Conference on Computer Vision and Pattern Recognition},
  pages 350--359, 2018.

\bibitem{hartley2003multiple}
R.~Hartley and A.~Zisserman.
\newblock {\em Multiple view geometry in computer vision}.
\newblock Cambridge university press, 2003.

\bibitem{jin2019multi}
S.~Jin, W.~Liu, W.~Ouyang, and C.~Qian.
\newblock Multi-person articulated tracking with spatial and temporal
  embeddings.
\newblock In {\em IEEE Conference on Computer Vision and Pattern Recognition},
  pages 5664--5673, 2019.

\bibitem{kamble2019ball}
P.~R. Kamble, A.~G. Keskar, and K.~M. Bhurchandi.
\newblock Ball tracking in sports: a survey.
\newblock {\em Artificial Intelligence Review}, 52(3):1655--1705, 2019.

\bibitem{kellnhofer18gaze360}
P.~Kellnhofer, A.~Recasens, S.~Stent, W.~Matusik, and A.~Torralba.
\newblock Gaze360: Physically unconstrained gaze estimation in the wild.
\newblock In {\em IEEE International Conference on Computer Vision (ICCV)},
  2019.

\bibitem{maksai2016players}
A.~Maksai, X.~Wang, and P.~Fua.
\newblock What players do with the ball: A physically constrained interaction
  modeling.
\newblock In {\em IEEE conference on computer vision and pattern recognition},
  pages 972--981, 2016.

\bibitem{manafifard2017survey}
M.~Manafifard, H.~Ebadi, and H.~A. Moghaddam.
\newblock A survey on player tracking in soccer videos.
\newblock {\em Computer Vision and Image Understanding}, 159:19--46, 2017.

\bibitem{ramakrishna2014pose}
V.~Ramakrishna, D.~Munoz, M.~Hebert, J.~A. Bagnell, and Y.~Sheikh.
\newblock Pose machines: Articulated pose estimation via inference machines.
\newblock In {\em European Conference on Computer Vision}, pages 33--47.
  Springer, 2014.

\bibitem{ran2019robust}
N.~Ran, L.~Kong, Y.~Wang, and Q.~Liu.
\newblock A robust multi-athlete tracking algorithm by exploiting discriminant
  features and long-term dependencies.
\newblock In {\em International Conference on Multimedia Modeling}, pages
  411--423. Springer, 2019.

\bibitem{seidl2018bhostgusters}
T.~Seidl, A.~Cherukumudi, A.~Hartnett, P.~Carr, and P.~Lucey.
\newblock Bhostgusters: Realtime interactive play sketching with synthesized
  {NBA} defenses.
\newblock In {\em MIT Sloan Sports Analytics Conference}, 2018.

\bibitem{thomas2017computer}
G.~Thomas, R.~Gade, T.~B. Moeslund, P.~Carr, and A.~Hilton.
\newblock Computer vision for sports: Current applications and research topics.
\newblock {\em Computer Vision and Image Understanding}, 159:3--18, 2017.

\bibitem{wang2019unsupervised}
N.~Wang, Y.~Song, C.~Ma, W.~Zhou, W.~Liu, and H.~Li.
\newblock Unsupervised deep tracking.
\newblock In {\em IEEE Conference on Computer Vision and Pattern Recognition},
  pages 1308--1317, 2019.

\bibitem{wang2019learning}
X.~Wang, A.~Jabri, and A.~A. Efros.
\newblock Learning correspondence from the cycle-consistency of time.
\newblock In {\em IEEE Conference on Computer Vision and Pattern Recognition},
  pages 2566--2576, 2019.

\bibitem{wei2016convolutional}
S.-E. Wei, V.~Ramakrishna, T.~Kanade, and Y.~Sheikh.
\newblock Convolutional pose machines.
\newblock In {\em IEEE Conference on Computer Vision and Pattern Recognition},
  pages 4724--4732, 2016.

\bibitem{zhang2018residual}
Y.~Zhang, Y.~Tian, Y.~Kong, B.~Zhong, and Y.~Fu.
\newblock Residual dense network for image super-resolution.
\newblock In {\em IEEE Conference on Computer Vision and Pattern Recognition},
  pages 2472--2481, 2018.

\end{thebibliography}
\end{document}